\documentclass{article} 
\usepackage{iclr2027_conference,times}

\usepackage{amsmath,amsfonts,bm}

\def\eqref#1{equation~\ref{#1}}

\def\1{\bm{1}}

\DeclareMathAlphabet{\mathsfit}{\encodingdefault}{\sfdefault}{m}{sl}
\SetMathAlphabet{\mathsfit}{bold}{\encodingdefault}{\sfdefault}{bx}{n}

\usepackage{hyperref}
\usepackage{url}
\usepackage{comment}
\usepackage{booktabs}
\usepackage{graphicx}
\usepackage{float}
\usepackage{flafter}
\usepackage{subcaption}
\usepackage{multirow}

\usepackage{algorithm}
\usepackage{algpseudocode}

\usepackage[section]{placeins}

\usepackage{xcolor}
\definecolor{highlightcolor}{RGB}{200, 0, 200}

\title{Cautious Conflict-Free Inverse Gradient Methods}

\iclrfinalcopy

\author{Damian Hattler, Qiang Liu, Nils Thuerey \\
School of Computation, Information and Technology\\
Technical University of Munich\\
Garching, DE 85748 \\
\texttt{damian.hattler@tum.de} \\}
\begin{document}

\maketitle

\begin{abstract}
Partial differential equations (PDEs) are indispensable tools for modeling complex phenomena in physics and engineering. Physics-Informed Neural Networks (PINNs) have emerged as a powerful alternative to traditional numerical solvers by directly incorporating physical constraints into neural network loss functions. Despite their advantages, PINN training requires multi-objective optimization, which often leads to significant gradient conflicts among boundary, initial, and residual loss terms. The Conflict-Free Inverse Gradient (ConFIG) method addresses these conflicts; however, integrating ConFIG with adaptive optimizers such as Adam introduces a critical vulnerability: the conflict-free direction generated by ConFIG is further processed by the optimizer, and the resulting weight update may no longer be conflict-free with respect to the task gradients.
To address this issue, Cautious ConFIG is proposed as a concise three-line extension to the ConFIG algorithm that empirically restores the conflict-freeness guarantee prior to the weight update. The method is evaluated on three PINN benchmarks: the 1D Viscous Burgers' equation, the 1D Nonlinear Schrödinger equation, and the 2D Kovasznay flow. Experimental results demonstrate consistent performance improvements over the standard ConFIG baseline. Additionally, the stabilizing effect of cautious masking and its reaction to conflict severity, task number, and optimizer lag are investigated using a simple toy system. The code is available at \url{https://github.com/damiha/CautiousConFIG}
\end{abstract}

\section{Introduction}

Partial differential equations (PDEs) are widely applied in physics \cite{sommerfeld1949partial}, chemistry \cite{jamwal2025applications}, engineering \cite{ames1965nonlinear}, and finance \cite{duffy2013finite} to model complex phenomena that evolve over space and time. These equations implicitly define a function of several variables, and solving a PDE with given initial and boundary conditions involves determining this function.
Traditionally, PDEs and their initial boundary value (IBV) problems
are solved using numerical methods that discretize the domain
into a finite set of subregions. The solution function is then represented as a set of values corresponding to these subregions. For example, the PDE can be reformulated as a time-stepping rule by approximating partial derivatives with finite differences.
Discretization may introduce artifacts, particularly for complex geometries, and the number of subregions can increase exponentially with the number of dimensions, a phenomenon known as the curse of dimensionality \cite{bellman2015adaptive}.
Physics-informed neural networks (PINNs) \cite{raissi2019physics}
have gained significant attention in recent years by directly parameterizing the solution function with a neural network. This approach mitigates discretization errors and alleviates the curse of dimensionality. As a mesh-free method, it is well-suited for complex geometries. Furthermore, formulating IBV problems as neural network training enables the utilization of recent advances in deep learning architectures and optimization techniques.
However, PINNs also present limitations, as deep learning methods are less well-understood regarding error analysis and generalization compared to classical numerical approaches. While numerical methods reliably yield solutions when conditions such as the Courant-Friedrichs-Lewy criterion \cite{courant1952solution} are satisfied, training a neural network in a high-dimensional loss landscape can be unpredictable.
One notable challenge in PINN training is gradient interference. Each IBV condition is formulated as a separate loss function, and the resulting gradients can interact adversely, severely hindering convergence.
The ConFIG method \cite{liu2025config} was developed to address the gradient interference problem by computing an update direction that is conflict-free with respect to all task gradients.
However, the ConFIG direction typically serves as an input to another optimizer, such as Adam \cite{kingma2014adam}, and the resulting update direction may not remain conflict-free with respect to the task gradients.
Consequently, the conflict-freeness property is often lost before the weight update is applied.
Our contributions are as follows:
\begin{itemize}
    \item We introduce Cautious ConFIG, a simple three-line extension compatible with any ConFIG-optimizer combination. It incurs negligible computational overhead, is easily parallelizable on the GPU, and effectively restores ConFIG's mathematical properties.

    \item We evaluate Cautious ConFIG on three challenging PINN benchmarks: the 1D viscous Burgers equation, the 1D nonlinear Schrödinger equation, and the 2D Kovasznay flow, observing substantial performance improvements over the baseline (19\% to 28\% relative improvement).

    \item We analyze the effectiveness of Cautious ConFIG by monitoring key metrics during PINN training and present a computational model demonstrating that Cautious ConFIG produces update directions that align more closely with task gradients under conditions of severe conflict, numerous tasks, and significant optimizer lag.
\end{itemize}

\section{Background}

Physical quantities, such as the velocity or pressure of a fluid,
are defined over a spatial domain $\Omega$ and evolve over the time interval $[0, T]$. In this context, only scalar quantities are considered. The objective is to determine a function $u: \Omega \times [0, T] \rightarrow \mathbb{R}$ that satisfies an \textbf{initial boundary value (IBV) problem}:
\begin{align*}
\mathcal{N}[u](x, t) := N[u](x, t) + f(x, t) &= 0, x \in \Omega, t \in (0, T] \\
\mathcal{B}[u](x, t) := B[u](x, t) + g(x, t) &= 0, x \in \partial \Omega, t \in (0, T] \\
\mathcal{I}[u] := u(x, 0) + h(x, 0) &= 0, x  \in \overline{\Omega}
\end{align*}
The initial values of $u$ are specified by the initial operator $\mathcal{I}$, while boundary values and their derivatives are defined using the boundary operator $\mathcal{B}$. The partial differential equation (PDE) itself constitutes only a component of the PDE operator $\mathcal{N}$.
IBV problems are commonly addressed using traditional numerical schemes, such as finite-difference or finite-element methods. In this work, we employ \textbf{physics-informed neural networks (PINNs)} \cite{raissi2019physics} to solve these problems. The conditions, represented by the operators $\mathcal{I}$, $\mathcal{B}$, and $\mathcal{N}$, are reformulated as loss functions. We denote the number of loss functions as $m$. The solution field $u$ is approximated by $u_\theta$, a neural network parameterized by $\theta \in \mathbb{R}^d$:
\begin{align*}
    \mathcal{L}_{init}(\theta) &= \mathbb{E}_{(x, t) \sim \mathcal{D}_{init}} \left[ (\mathcal{I}[u_{\theta}](x, t))^2\right] \\
    \mathcal{L}_{bound}(\theta) &= \mathbb{E}_{(x, t) \sim \mathcal{D}_{bound}} \left[ (\mathcal{B}[u_{\theta}](x, t))^2\right] \\
    \mathcal{L}_{pde}(\theta) &= \mathbb{E}_{(x, t) \sim \mathcal{D}_{pde}} \left[ (\mathcal{N}[u_{\theta}](x, t))^2\right]
\end{align*}
The $m$ loss functions are differentiable with respect to $\theta$ because the neural network is fully differentiable with respect to $\theta$ and $(x, t)$. Consequently, these loss functions can be computed using an autograd engine such as PyTorch \cite{paszke2019pytorch}. As a result, solving an IBV problem is reformulated as an optimization problem involving $m$ loss functions.
Minimizing $m$ loss functions falls within the domain of \textbf{multi-task learning (MTL)}. When all loss functions are differentiable with respect to $\theta$, \textbf{gradient-based optimization techniques} can be applied. A common approach is to form a linear combination of the loss functions, resulting in a single scalar loss function $\mathcal{L}(\theta) = \sum_{i = 1}^m w_i \mathcal{L}(\theta)$. Classic optimization methods, such as gradient descent, can then be used to minimize $\mathcal{L}$. However, these \textbf{scalarization-based}
approaches are susceptible to \textbf{gradient conflicts}.
Gradient conflicts as a source of problems in MTL training were first identified by \cite{yu2020gradient}. Two gradients $g_i$ and $g_j$ (corresponding to the loss functions $\mathcal{L}_i$ and $\mathcal{L}_j$
respectively) are said to exhibit a gradient conflict when $\langle g_i, g_j \rangle < 0$. If one gradient is significantly larger than the other, the resulting update direction $g$, which is a linear combination of $g_i$ and $g_j$, may conflict with one of them, indicated by a negative dot product with $g_i$ or $g_j$. Updating parameters in the direction $-\eta \; g$ can therefore minimize one loss while increasing another, provided the learning rate $\eta$ is sufficiently small. This outcome is undesirable. Furthermore, if the gradients point in opposite directions, they may cancel, thereby slowing down learning. The susceptibility of PINNs to gradient conflicts is demonstrated in \cite{liu2025config}.
At a specific location $\theta$ in parameter space, the gradients $g_1, ..., g_m$ are fixed, assuming the input $(x, t)$ is not randomized. Each gradient $g_i$ defines a set of directions that are conflict-free with respect to it. This is referred to as the half-space $\mathcal{H}_i = \{v: \langle v, g_i \rangle \geq 0 \}$. The task gradient $g_i$ acts as the normal vector to a hyperplane, with $\mathcal{H}_i$ representing one side of this hyperplane. The intersection of all gradient half-spaces defines the set of \textbf{common-descent} or \textbf{conflict-free directions} $\mathcal{C} = \bigcap_{i = 1}^m \mathcal{H}_i$.
Updating the weights in the direction $-\eta \; v$, where $v \in \mathcal{C}$, ensures that all $m$ losses decrease, provided $\eta$ is sufficiently small.
A specific optimization direction within $\mathcal{C}$ is defined by the \textbf{Conflict-Free Inverse Gradient (ConFIG)} \cite{liu2025config} method. The optimization direction $g_{\text{ConFIG}}$ is constructed to have positive and equal cosine similarities with all gradients $g_i$. This direction is given by the following equations:
\begin{align*}
    g_u &= \mathcal{U}((\left[ \mathcal{U}(g_1), …, \mathcal{U}(g_m) \right]^T)^{\dagger} \textbf{1}_m) \\
g_{\text{ConFIG}} &= \left(\sum_{i = 1}^m g_i^T g_u \right) g_u \\
\end{align*}
The normalization $\mathcal{U}$ scales all task gradients to unit $L_2$ norm, ensuring that the direction of $g_{\text{ConFIG}}$ is independent of the magnitudes of $g_1, ..., g_m$. The scalar factor $\sum_{i = 1}^m \langle g_i, g_u \rangle$ functions as a conflict-aware learning rate.
Previous discussion has focused exclusively on \textbf{spatial conflicts}, which are intrinsic to $g_1, ..., g_m$ and are associated with specific regions in parameter space. However, the use of momentum-based optimizers such as Adam \cite{kingma2014adam} can also introduce conflicts. When the current gradient is provided as input to such an optimizer, it updates the optimizer's internal state. The optimizer's output $u$ combines the current gradient $g$ with state variables that accumulate the history of past gradients throughout the optimization process.
The weights are updated with $-\eta \, u$, but it is possible that $\langle u, g\rangle < 0$. In this case, the current gradient $g$ does not sufficiently influence the optimizer state, resulting in a weight update that increases the loss. This clash between the present (represented by $g$) and the past (represented by $u$) is referred to as a \textbf{temporal conflict}.
Temporal conflicts can be addressed by employing a \textbf{cautious update} \cite{liang2026cautious}, defined as follows:
\begin{align*}
    \phi &= \mathbf{1}\{ u \odot g > 0\} \\
    \alpha &= \frac{d}{\lVert \phi \rVert_1 + 1} \\
    u_{cautious} &= \alpha (u \odot \phi)
\end{align*}
Masking updates where $g$ and $u$ disagree ensures
that $\langle u_{cautious}, g \rangle \geq 0$. The cautious rescaling factor $\alpha$ maintains $u_{cautious}$ at approximately the same $L_1$ norm as $u$.

\section{Related Work}

The proposed method, Cautious ConFIG, is directly derived from
the ConFIG method \cite{liu2025config} and cautious optimizers \cite{courant1952solution}. ConFIG addresses spatial conflicts, whereas cautious updates resolve temporal conflicts introduced by adaptive optimizers.
Several alternatives to ConFIG exist for resolving spatial conflicts. The PCGrad method \cite{yu2020gradient} subtracts the projection of one gradient from another to eliminate pairwise gradient conflicts. This process removes shared directional components, thereby preventing gradient interference. However, because these subtractions are performed randomly before summing the cleaned gradients, PCGrad cannot guarantee, as ConFIG does, that $g_{\text{PCGrad}}$ is conflict-free with respect to the input gradients. Nash-MTL \cite{navon2022multi} formulates multi-task learning as a game and derives a conflict-free direction using the Nash bargaining solution. CAGrad \cite{liu2021conflict} identifies a conflict-free direction that maximizes the minimum dot product with all task gradients while remaining close to the average gradient $g_0 = \frac{1}{m} \sum_{i = 1}^m g_i$.
Early methods for resolving temporal conflicts include RProp \cite{riedmiller1993direct}, which compares gradients $g_t$ and $g_{t - 1}$ elementwise based on their sign. If the signs agree, the parameter-specific learning rate is increased multiplicatively; otherwise, the learning rate is decreased. Magma \cite{joo2026surprising} represents another alternative to cautious optimizers.
In Magma, the cosine similarity between $u$ and $g$ is computed for each parameter group, such as a weight matrix in a neural network. The number of gradient updates skipped depends on the cosine similarity. The $L_1$ norm lost due to masking is redistributed to the remaining coordinates, analogous to the approach used in cautious optimizers.
The work most closely related to Cautious ConFIG is Gradient Update Mismatch (GUM) \cite{xiao2026gradient}, which was developed concurrently; therefore, a direct comparison is not provided. GUM addresses the same issue, namely conflicts arising from introducing an optimizer after ConFIG, referred to as gradient update mismatch. The authors specifically analyze the Adam and ConFIG combination and evaluate their approach on the same IBV problems (Burgers, Schrödinger, Kovasznay).
However, notable differences exist between their proposed solution, Gradient Update Alignment (GUA), and Cautious ConFIG regarding geometric approach and solution complexity. In GUA, the optimizer output $u$ is projected onto $\mathcal{C}$, referred to as the conflict-free cone.
This projection is achieved by solving a quadratic program, where a custom metric $M_t$, such as the Adam metric, may be employed.
In contrast, Cautious ConFIG projects $u$ onto the orthant defined by $g_{\text{ConFIG}}$, resulting in a solution that is invariant to the choice of metric. While solving a quadratic program in GUA requires an active set solver, which is challenging to parallelize on GPUs due to branching logic, the Cautious ConFIG approach is easily parallelizable, involving only the application of a binary mask. Furthermore, Cautious ConFIG is significantly simpler to implement, requiring only three lines of code compared to more than 120 lines for the GUA projection logic. Additionally, GUA introduces a state-alignment procedure that updates the optimizer state after projection using an exponential moving average update of Adam's $m$ and $v$, introducing two new hyperparameters, $\rho_m$ and $\rho_v$, that require tuning. Cautious ConFIG does not introduce any new hyperparameters.

\section{Proposed Approach}

\subsection{Problem}

ConFIG computes a direction denoted as $g_{\text{ConFIG}}$.
This direction satisfies $\langle g_{\text{ConFIG}}, g_i \rangle \geq 0$ for all task gradients $g_i$. In other words, ConFIG generates a direction that is conflict-free with respect to all task gradients. If the ConFIG direction is used for the weight update $\theta \leftarrow \theta - \eta g_{\text{ConFIG}}$,
then all losses $\mathcal{L}_i$ decrease after the optimization step, provided the learning rate is sufficiently small.
In practical applications, ConFIG serves as an input to another optimizer, such as Adam. Adam incorporates both momentum and adaptiveness, and both aspects
can lead to a loss of the conflict-freeness guarantee. Although a direction conflict-free with the task gradients is provided to Adam, the resulting direction may not retain this property. This output direction is then used in the weight update.
Adam updates the weights in the direction $-\eta \frac{u^{(t)}}{\sqrt{v^{(t)}} + \epsilon}$. When $v$ is set to $\mathbf{1}_d$,
adaptiveness is disabled. In this scenario, updating with $m^{(t)}$ can still be problematic because the current first moment consists of 90\% $m^{(t-1)}$ and only 10\% $g_{\text{ConFIG}}$ (for the default setting $\beta_1 = 0.9$).
If $m^{(t-1)}$ was not conflict-free with the task gradients,
$m^{(t)}$ might also lack conflict-freeness, as $g_{\text{ConFIG}}$ may not adjust the first moment sufficiently quickly. Consequently, the weights may move in a conflicting direction.
Adaptiveness can also result in the loss of the conflict-freeness property. When $\beta_1 = 0$, the update is performed as $\theta \leftarrow \theta - \eta \frac{ g_{\text{ConFIG}}}{\sqrt{v} + \epsilon}$. This corresponds to gradient descent with the Adam metric $M_t = \text{diag}(\sqrt{v} + \epsilon)$. The ConFIG algorithm ensures $\langle g_{\text{ConFIG}}, g_i \rangle_I \geq 0$ (conflict-freeness with respect to the Euclidean metric), but not $\langle g_{\text{ConFIG}}, g_i \rangle_{M_t} \geq 0$. As a result, the weights may again move in a conflicting direction.

\subsection{Cautious ConFIG}

\begin{figure}[htbp]
    \centering
    \includegraphics[width=0.5\linewidth]{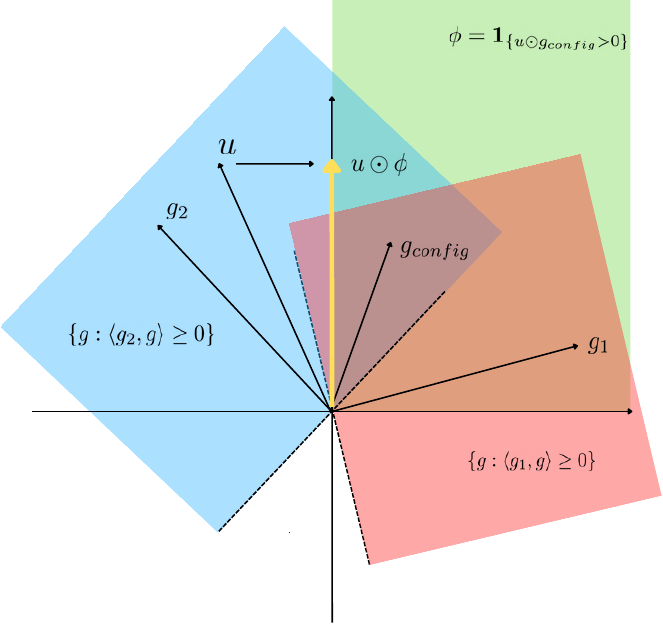}
    \caption{Cautious ConFIG projects the optimizer output $u$ onto the orthant given by $g_{\text{ConFIG}}$}
    \label{fig:orthant_projection}
\end{figure}

The optimizer can be regarded as a black box that transforms $g_{\text{ConFIG}}$ into $u^{(t)}$, which is subsequently used for the weight update. A key issue arises when the directions of $g_{\text{ConFIG}}$ and $u^{(t)}$ stronly diverge. When $u^{(t)}$ and $g_{\text{ConFIG}}$ exhibit low cosine similarity, $u^{(t)}$ and $g_i$ also tend to have low similarity, at least heuristically.
%
The objective is to transform $u^{(t)}$ into a direction $u_{modified}^{(t)}$ that maintains a positive dot product with all task gradients. This is achieved through a \emph{soft alignment strategy}, which brings $u_{modified}^{(t)}$ closer to $g_{\text{ConFIG}}$. Since $g_{\text{ConFIG}}$ exhibits maximally large cosine similarities with all task gradients as the maximum angular margin solution, aligning $u_{modified}^{(t)}$ with $g_{\text{ConFIG}}$ is expected to increase its cosine similarity with all task gradients. This hypothesis is supported by the experimental results shown in Figure [REF].
\begin{figure}[htbp]
    \centering
    \includegraphics[width=0.70\linewidth]{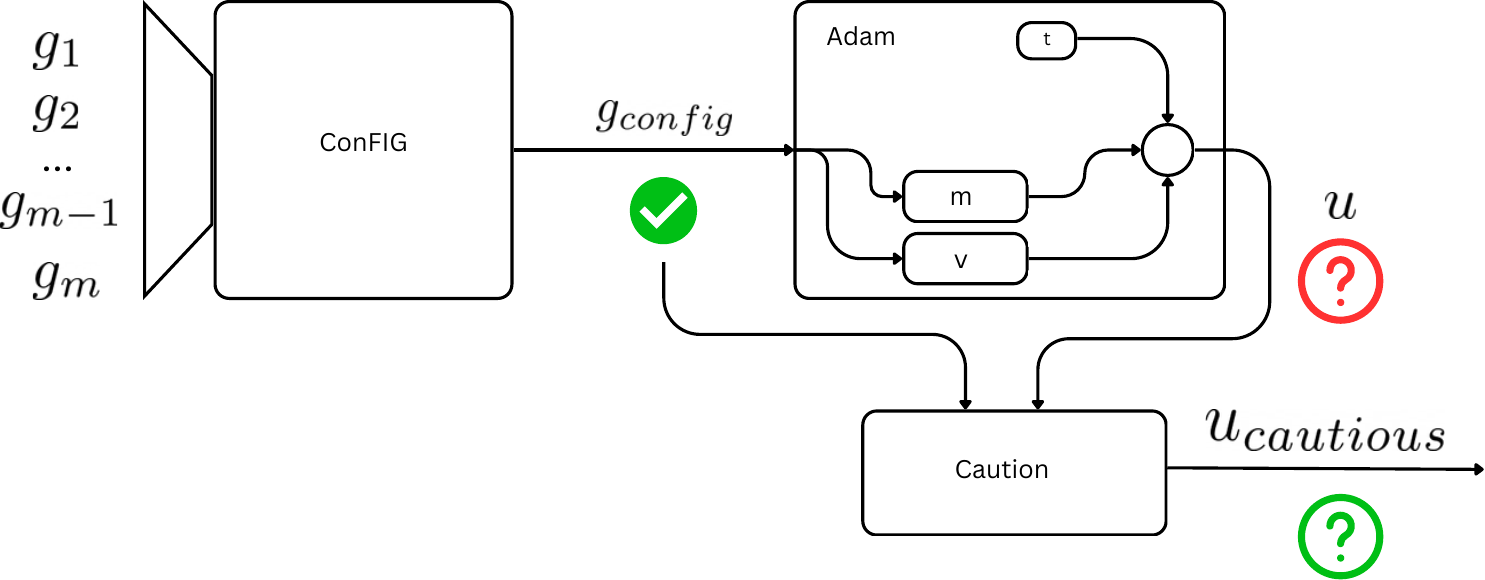}
    \caption{The output of "Adam + ConFIG" is post-processed using a cautious update to tie it closer to the conflict-free input.}
    \label{fig:pipeline_with_caution}
\end{figure}
%
Cautious ConFIG constitutes a concise extension to the Adam + ConFIG combination (see Algorithm \ref{alg:cautious_config} for the pseudo code and Figure \ref{fig:pipeline_with_caution} for the pipeline). The optimizer output $u^{(t)}$ is adjusted towards $g_{\text{ConFIG}}$ through a cautious update, ensuring that $\text{Cos}(u_{cautious}^{(t)}, g_{\text{ConFIG}}) \geq \text{Cos}(u^{(t)}, g_{\text{ConFIG}})$ (see Proposition \ref{prop:cautious_alignment}). In this approach, $g_{\text{ConFIG}}$ serves as a geometric anchor, selecting an orthant $\mathbb{O}_{g_\text{ConFIG}} = \{ v: \text{sign}(v) = \text{sign}(g_{\text{ConFIG}})\}$ for orthogonal projection, followed by rescaling to preserve the $L_1$ norm. The geometric interpretation is depicted in Figure \ref{fig:orthant_projection}. Because $g_{\text{ConFIG}}$ is centrally located within the conflict-free cone $\mathcal{C}_t$, the selected orthant typically overlaps substantially with this cone. Projecting onto the orthant frequently results in $u_{cautious}^{(t)}$ being a common descent direction.
%
Since multiplying $u^{(t)}$ with a binary mask $\phi^{(t)} = \mathbf{1}\{ u^{(t)} \odot g^{(t)} > 0\}$
ensures $\langle u^{(t)}, g^{(t)} \rangle \geq 0$ (see Proposition \ref{prop:cautious_pos_dot_prod}), we could turn $u^{(t)}$ into a conflict-free direction
just by masking alone. We would calculate $\phi_i^{(t)} = \mathbf{1}\{ u^{(t)} \odot g_i^{(t)} > 0\}$ and then
multiply $u^{(t)}$ elementwise with $\phi^{(t)} = \Pi_{i = 1}^m \phi_i^{(t)}$. This process, known as \emph{hard alignment}, aligns $u^{(t)}$ directly with the task gradients $g_i$. In this scenario, $g_{\text{ConFIG}}$ becomes unnecessary because conflict-freeness is achieved solely through masking.
The downside of this is that the number of parameters updated decreases exponentially in $m$ (all $m$ gradient components have to agree on the same sign, assuming each sign occurs with probability $p = 0.5$, the number of updated parameters is proportional to $2^{-m}$). This exponential reduction in updated parameters
negates the benefits of multi-task learning and leads to rapid destabilization, as the exponentially reduced set of updated parameters receives a proportionally high learning rate.
%
Therefore, to eliminate dependence on $m$, the number of tasks, cautious masking should be applied only once to enforce alignment of $u^{(t)}$ with a single target direction that summarizes all $m$ task gradients. This summary direction is the ConFIG direction.

\begin{algorithm}[htbp]
\caption{Cautious ConFIG}
\label{alg:cautious_config}
\begin{algorithmic}[1]
\State $m^{(0)} \leftarrow 0 \in \mathbb{R}^d$ \Comment First momentum
\State $v^{(0)} \leftarrow 0 \in \mathbb{R}^d$ \Comment Second momemtum

\For{$t = 1 \text{ to n\_epochs}$}

    \State $[g_1^{(t)}, ..., g_m^{(t)}] \leftarrow [\nabla_{\theta} L_1(\theta), ..., \nabla_{\theta} L_m(\theta)]$ \Comment m backward passes

    \State $g_u \leftarrow \mathcal{U}([\mathcal{U}(g_1), ..., \mathcal{U}(g_m)]^{-T} \mathbf{1}_m)$

    \State $g_{config} \leftarrow \left(\sum_{i = 1}^m g_i^T g_u \right) g_u$

    \State $m^{(t)} \leftarrow \beta_1 m^{(t - 1)} + (1 - \beta_1) g_{config}$

    \State $v^{(t)} \leftarrow \beta_2 v^{(t - 1)} + (1 - \beta_2) (g_{config} \odot g_{config})$

    \State $\hat{m}^{(t)} \leftarrow \frac{m^{(t)}}{1 - \beta_1^t}$ \Comment Debias first momentum

    \State $\hat{v}^{(t)} \leftarrow \frac{v^{(t)}}{1 - \beta_2^t}$ \Comment Debias second momentum

    \State $u^{(t)} \leftarrow \frac{\hat{m}^{(t)}}{\sqrt{\hat{v}^{(t)}} + \epsilon}$ \Comment Calculate Adam direction

    \State \textcolor{highlightcolor}{$\phi \leftarrow \mathbf{1}\{ u^{(t)} \odot g_{config} > 0 \}$ \Comment Binary mask}

    \State \textcolor{highlightcolor}{$\alpha \leftarrow \frac{d}{|\phi|_1 + 1}$ \Comment Cautious rescaling}

    \State \textcolor{highlightcolor}{$u^{(t)} \leftarrow \alpha \, (u^{(t)} \odot \phi)$}

    \State $\theta^{(t)} \leftarrow \theta^{(t - 1)} - \eta \cdot u^{(t)}$ \Comment Parameter update 
\EndFor
\end{algorithmic}
\end{algorithm}

\section{Experiments}
\begin{figure}[htbp]
    \centering
    \begin{subfigure}[b]{0.44\linewidth}
        \centering
        \includegraphics[width=\linewidth]{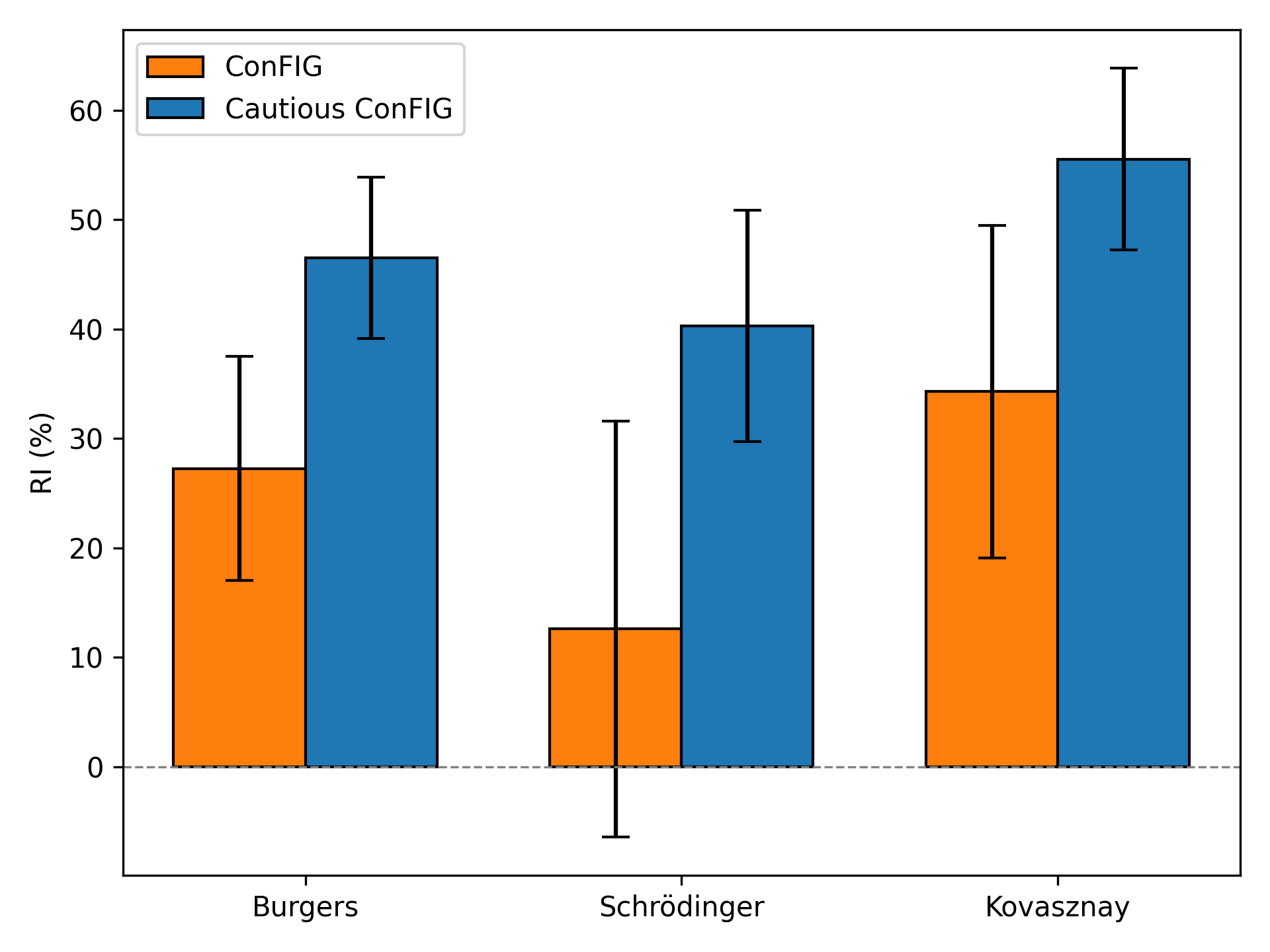}
        \caption{Baseline: Adam}
    \end{subfigure}%
    \hfill
    \begin{subfigure}[b]{0.48\linewidth}
        \centering
        \includegraphics[width=\linewidth]{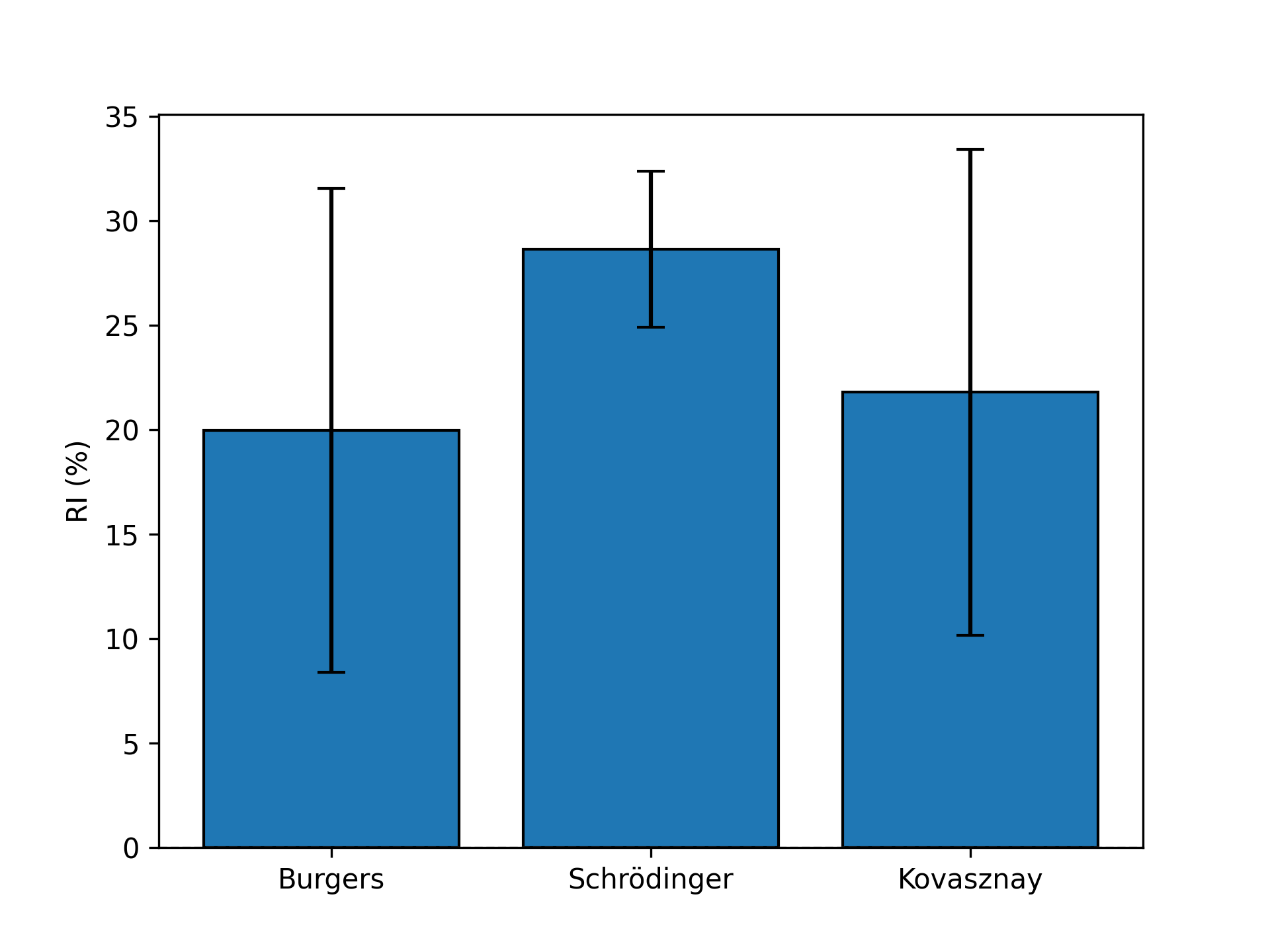}
        \caption{Baseline: Adam + ConFIG}
    \end{subfigure}
    \caption{Relative improvement to the baseline}
    \label{fig:relative_improvement}
\end{figure}
\begin{figure}
    \centering
    \begin{subfigure}[b]{0.32\linewidth}
        \centering
        \includegraphics[width=\linewidth]{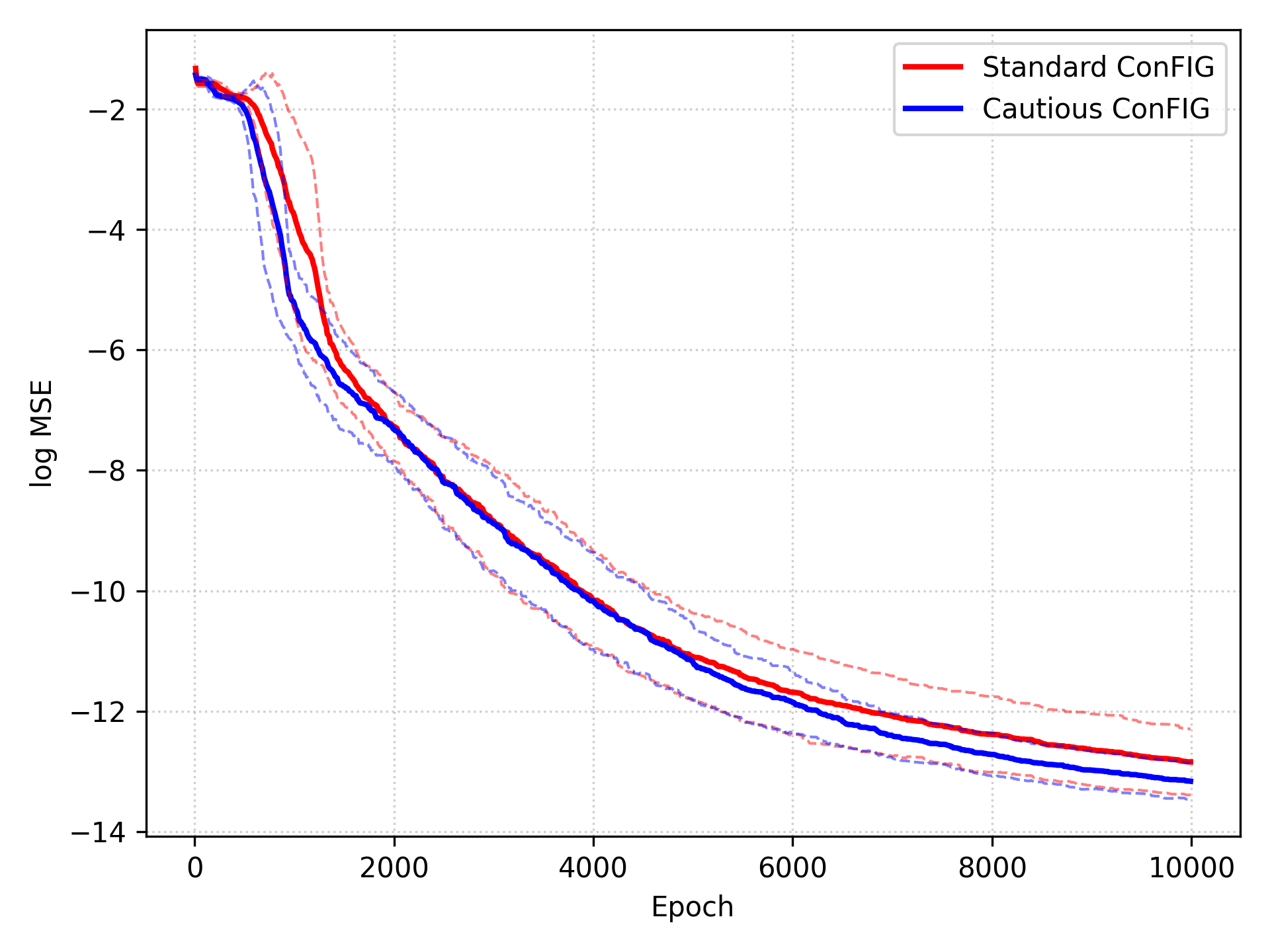}
        \caption{Kovasznay (two losses)}
    \end{subfigure}%
    \hfill 
    \begin{subfigure}[b]{0.32\linewidth}
        \centering
        \includegraphics[width=\linewidth]{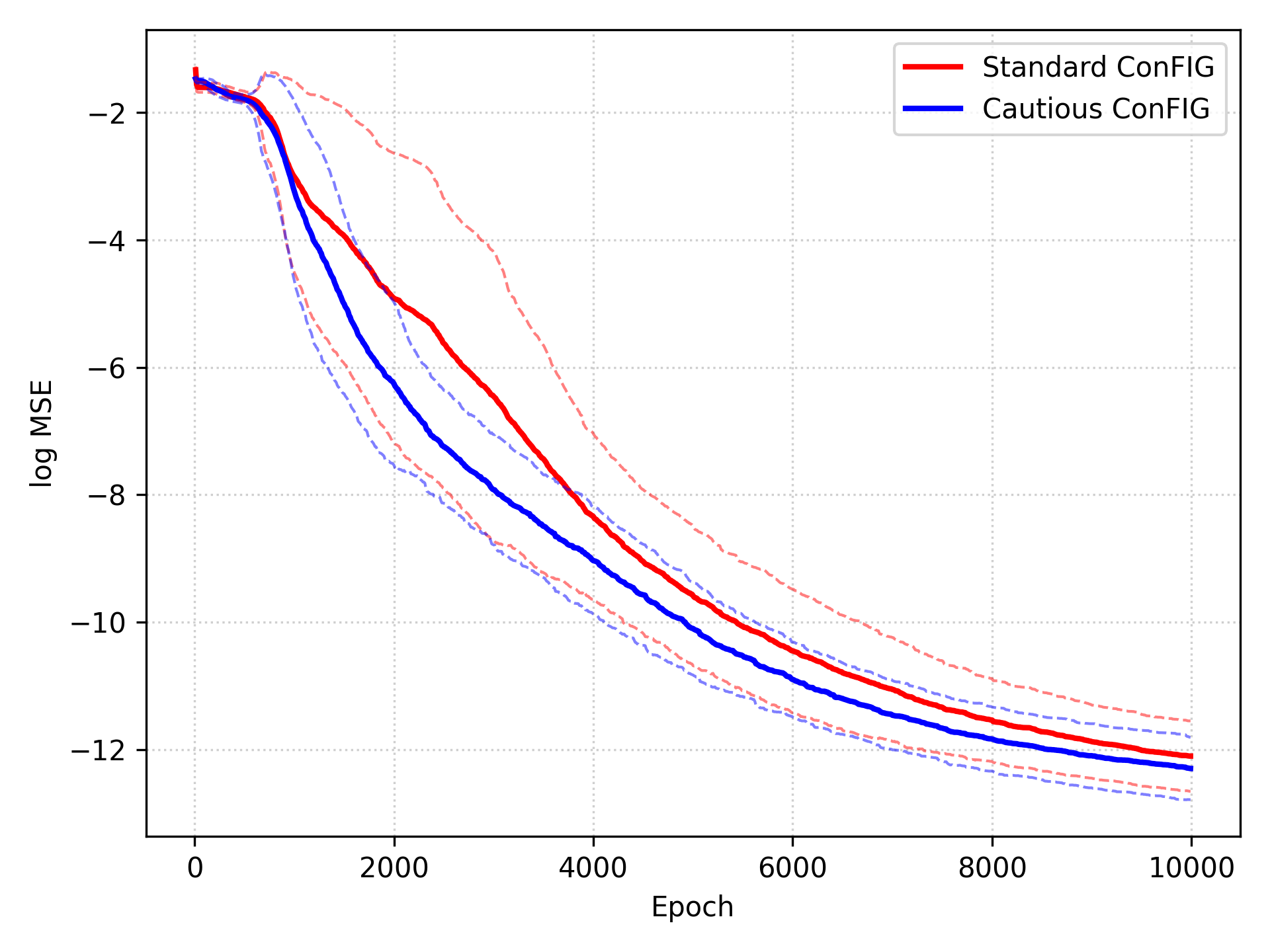}
        \caption{Kovasznay (three losses)}
    \end{subfigure}%
    \hfill
    \begin{subfigure}[b]{0.32\linewidth}
        \centering
        \includegraphics[width=\linewidth]{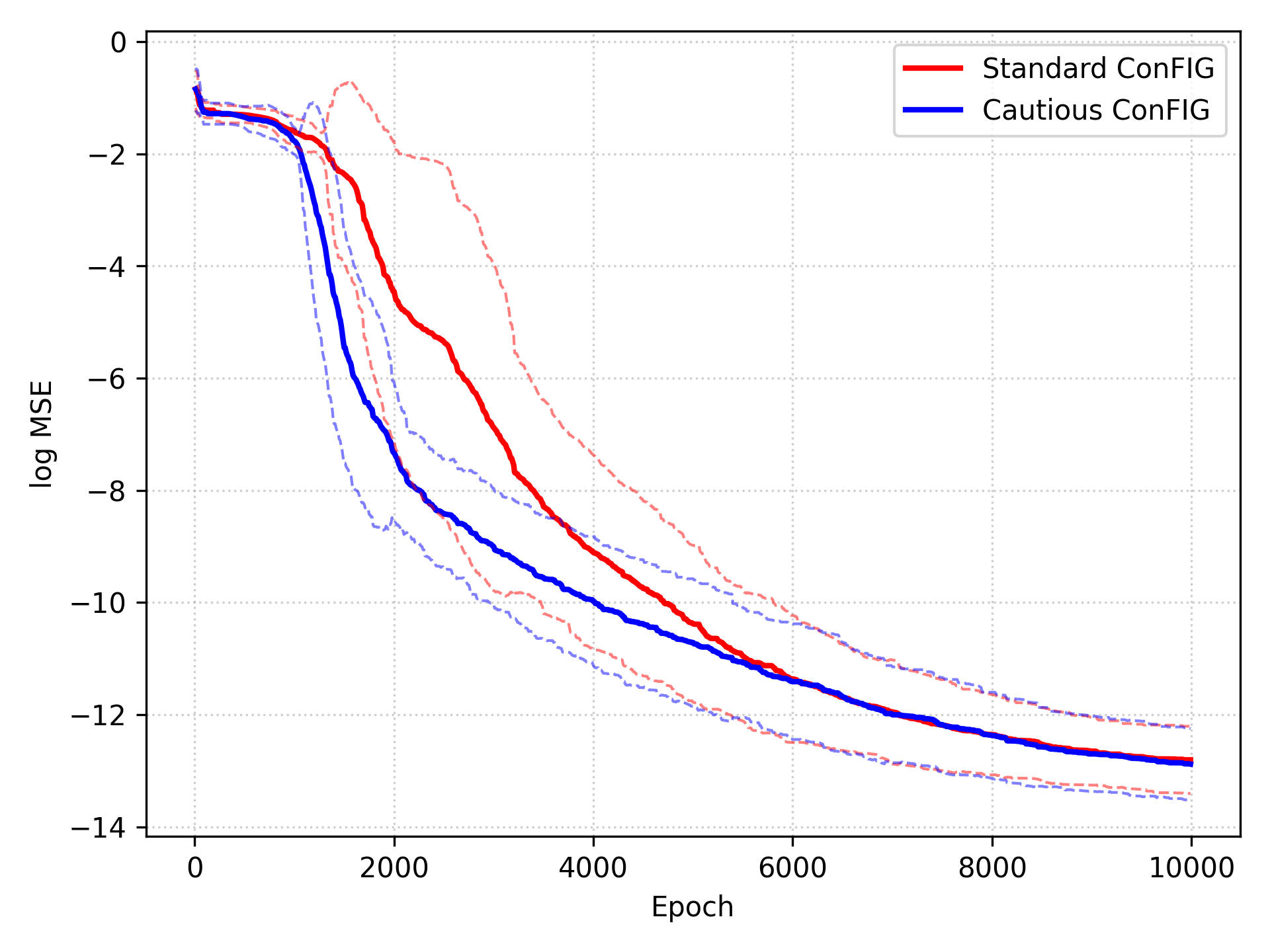}
        \caption{Kovasznay (six losses)}
    \end{subfigure}
    \caption{Cautious ConFIG converges faster than the Adam + ConFIG baseline. The performance gap widens with more tasks.}
    \label{fig:cautious_config_converges_faster}
\end{figure}

Cautious ConFIG is evaluated on three nonlinear PDEs: the 1D viscous Burgers equation, the 1D nonlinear Schrödinger equation, and the 2D Kovasznay flow. Definitions of the IBVs are provided in Appendix \ref{app:experimental details}. The minimum mean squared error (MSE) between the PINN prediction and the numerical solution achieved during training is used as the scalar performance metric. Figure \ref{fig:relative_improvement} presents these results.
The results indicate that Cautious ConFIG yields substantial relative improvements over both plain Adam, where gradient conflicts are present, and Adam + ConFIG, where the mathematical guarantees of ConFIG are not preserved at the optimizer output. The observed relative improvement ranges from 19\% to 28\% compared to Adam + ConFIG. Shared hyperparameters are listed in Table \ref{tab:shared_hyperparameters} and are consistent with those used in the original ConFIG study. However, PINN-specific hyperparameters, such as the number of sampling points for the domain boundary and interior, and the number of training epochs, differ from those in the original paper due to time and resource constraints. Consequently, the minimum MSE values are not directly comparable to those reported previously. The key finding is that the relative improvement over Adam + ConFIG remains substantial.
\begin{table}[htbp]
    \centering
    \caption{Mean Squared Error (MSE) and standard deviation over 10 random seeds for Viscous Burgers, Nonlinear Schrödinger, and Kovasznay flow.}
    \label{tab:optimizer_results_combined}
    \begin{tabular}{lccc}
        \toprule
        \multirow{2}{*}{\textbf{Optimizer}} & \multicolumn{3}{c}{\textbf{MSE}} \\
        \cmidrule(lr){2-4}
        & \textbf{Burgers} ($\times 10^{-4}$) & \textbf{Schrödinger} ($\times 10^{-4}$) & \textbf{Kovasznay} ($\times 10^{-6}$) \\
        \midrule
        Cautious ConFIG (ours) & \textbf{5.775 $\pm$ 1.193} & \textbf{2.096 $\pm$ 0.772} & \textbf{2.013 $\pm$ 0.757} \\
        Adam + ConFIG          & 8.367 $\pm$ 3.482 & 2.975 $\pm$ 1.121 & 3.098 $\pm$ 1.885 \\
        Adam + Caution         & 9.100 $\pm$ 4.926 & 3.876 $\pm$ 1.811 & 5.587 $\pm$ 2.463 \\
        Adam                   & 12.873 $\pm$ 6.119 & 4.171 $\pm$ 1.745 & 5.540 $\pm$ 2.547 \\
        \bottomrule
    \end{tabular}
\end{table}
Table \ref{tab:optimizer_results_combined} demonstrates that both the mean MSE and its standard deviation, measured over 10 random seeds, are reduced. Cautious ConFIG exhibits higher performance and can be trained with greater stability. Figure \ref{fig:cautious_config_converges_faster} presents three versions of the Kovasznay flow problem with varying numbers of tasks ($m \in \{2, 3, 4 \}$). Cautious ConFIG converges more rapidly in the initial stages, and the performance gap increases as the number of tasks grows. This advantage for larger $m$ will be further illustrated in the subsequent toy model.

\section{Analysis}

\begin{figure}
    \centering
    \begin{subfigure}[b]{\linewidth}
        \centering
        \includegraphics[width=\linewidth]{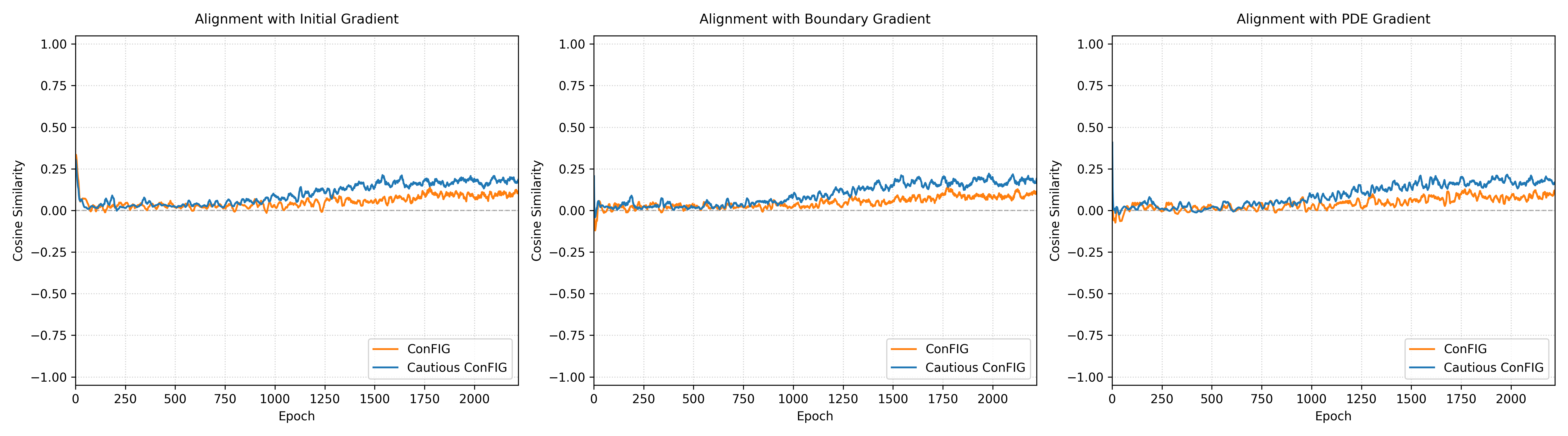}
        \caption{Burgers}
    \end{subfigure}
    
    \vspace{0cm}

    \begin{subfigure}[b]{\linewidth}
        \centering
        \includegraphics[width=\linewidth]{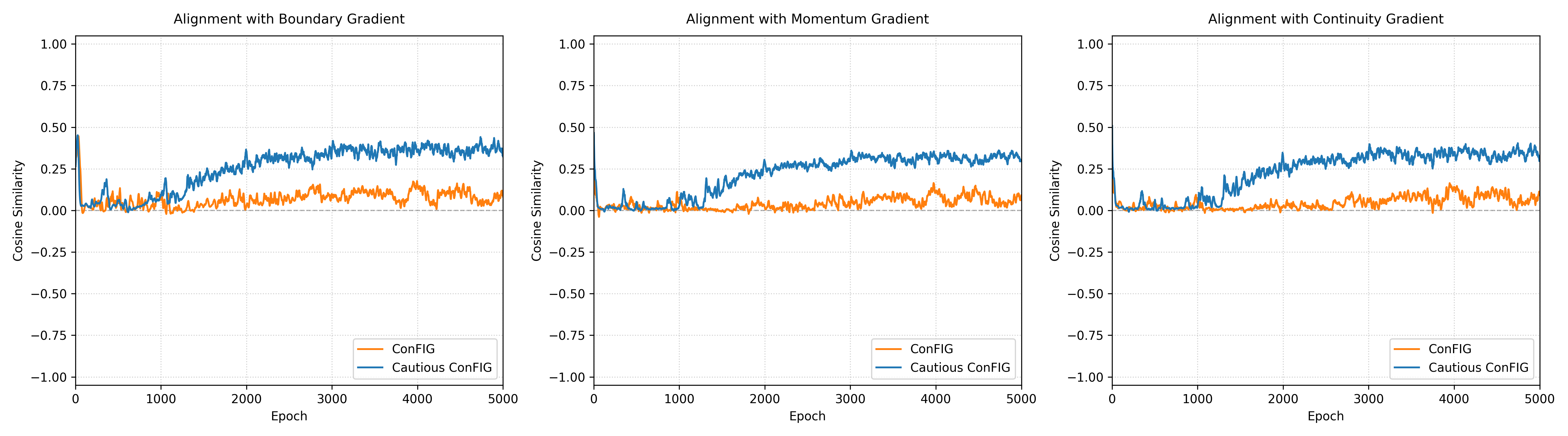}
        \caption{Kovasznay (three losses)}
    \end{subfigure}
    \caption{Cosine similarity between update direction and task gradients}
    \label{fig:update_task_alignment}
\end{figure}

\begin{figure}[htbp]
    \centering
    \begin{subfigure}[b]{\textwidth}
        \centering
        \includegraphics[width=\linewidth]{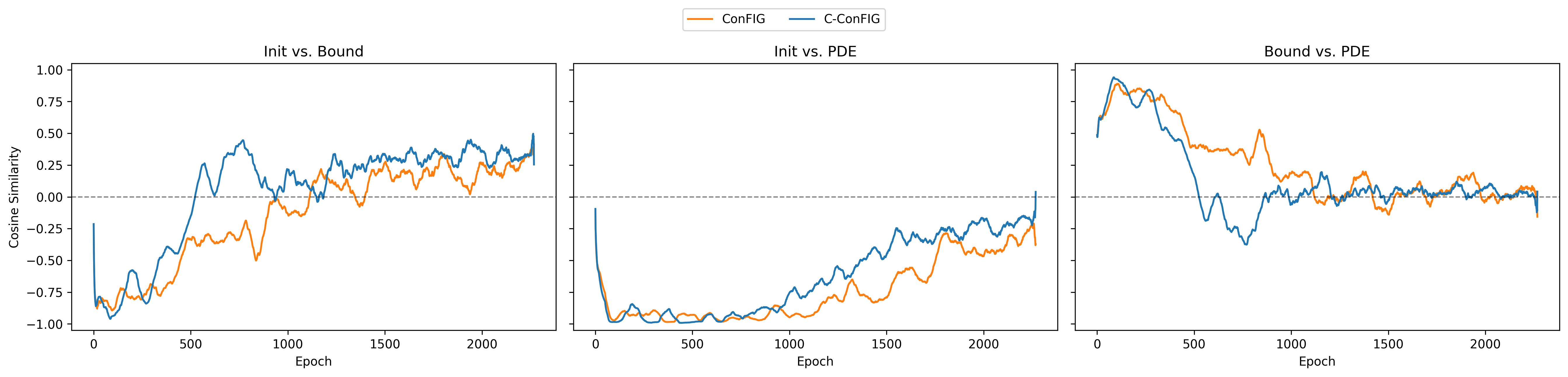}
        \caption{Burgers}
    \end{subfigure}
    
    \vspace{0.3cm}
    
    \begin{subfigure}[b]{\textwidth}
        \centering
        \includegraphics[width=\linewidth]{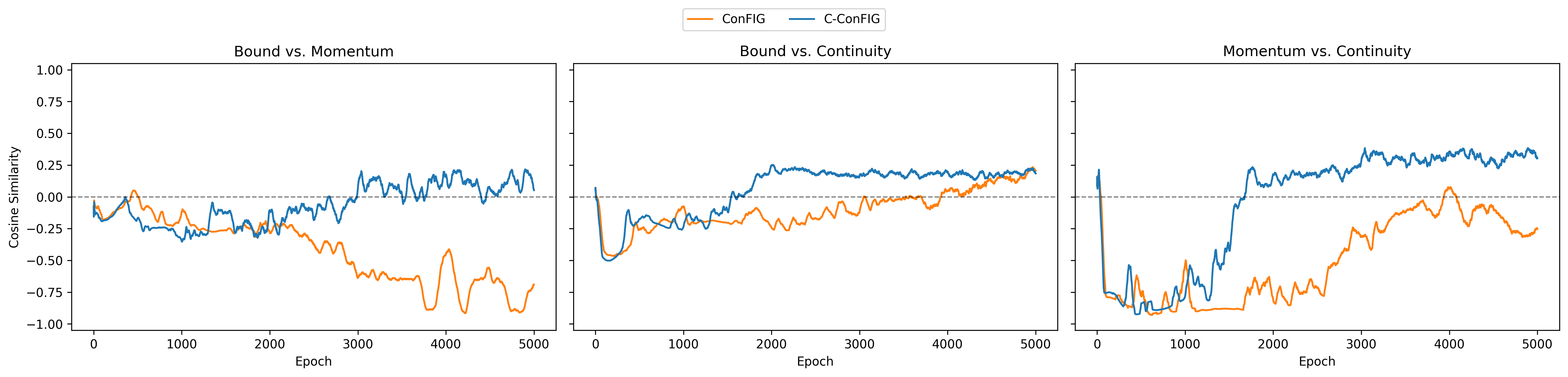}
        \caption{Kovasznay (three losses)}
    \end{subfigure}
    
    \caption{Task conflicts quantified via gradient cosine similarities during the initial training phase across different PINNs}
    \label{fig:burgers_kovasznay_task_conflicts_initial_phase}
\end{figure}

\begin{figure}[htbp]
    \centering
    \begin{subfigure}[b]{0.32\textwidth}
        \centering
        \includegraphics[width=\linewidth]{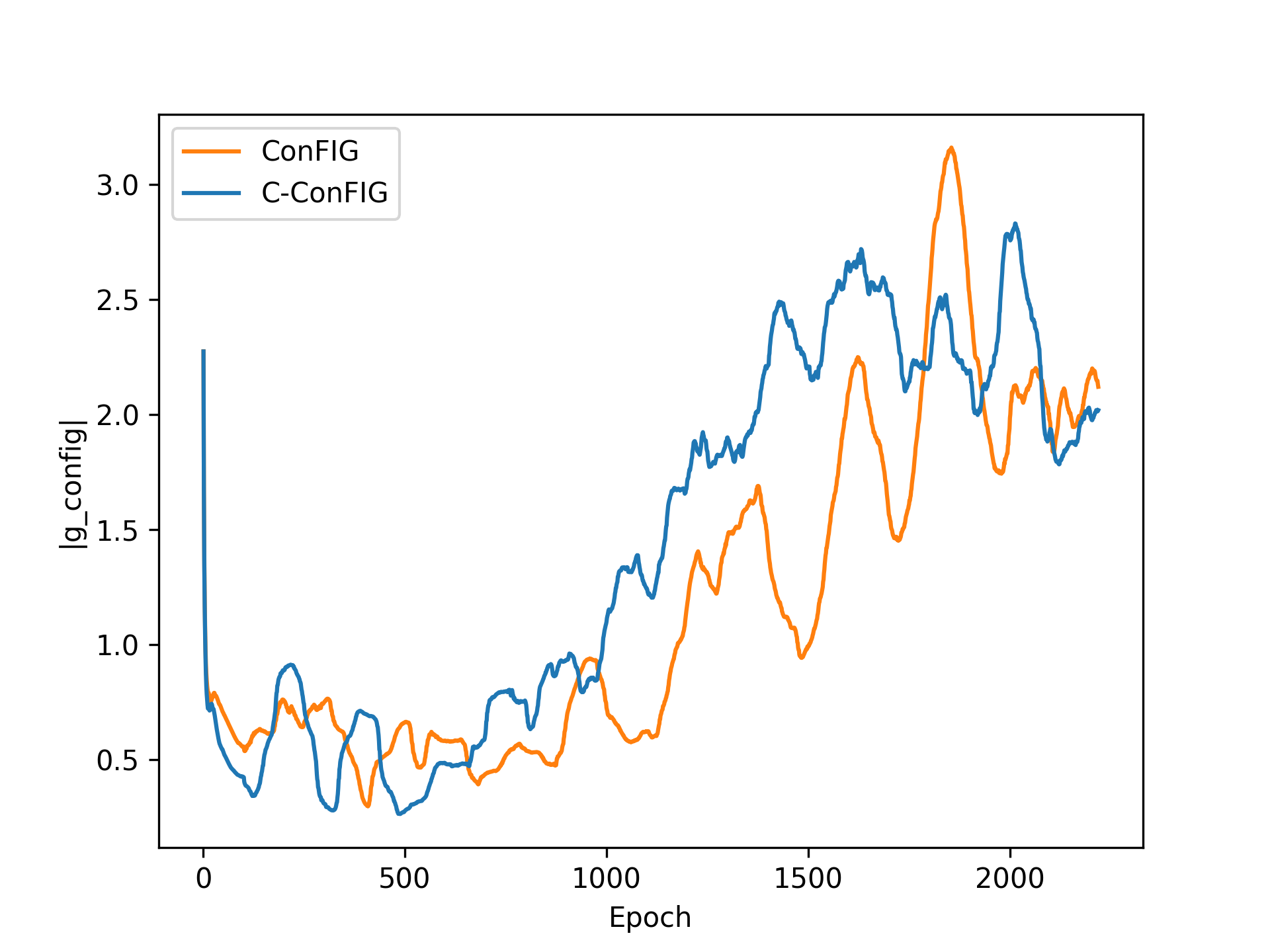}
        \caption{Burgers}
    \end{subfigure}
    \hfill
    \begin{subfigure}[b]{0.32\textwidth}
        \centering
        \includegraphics[width=\linewidth]{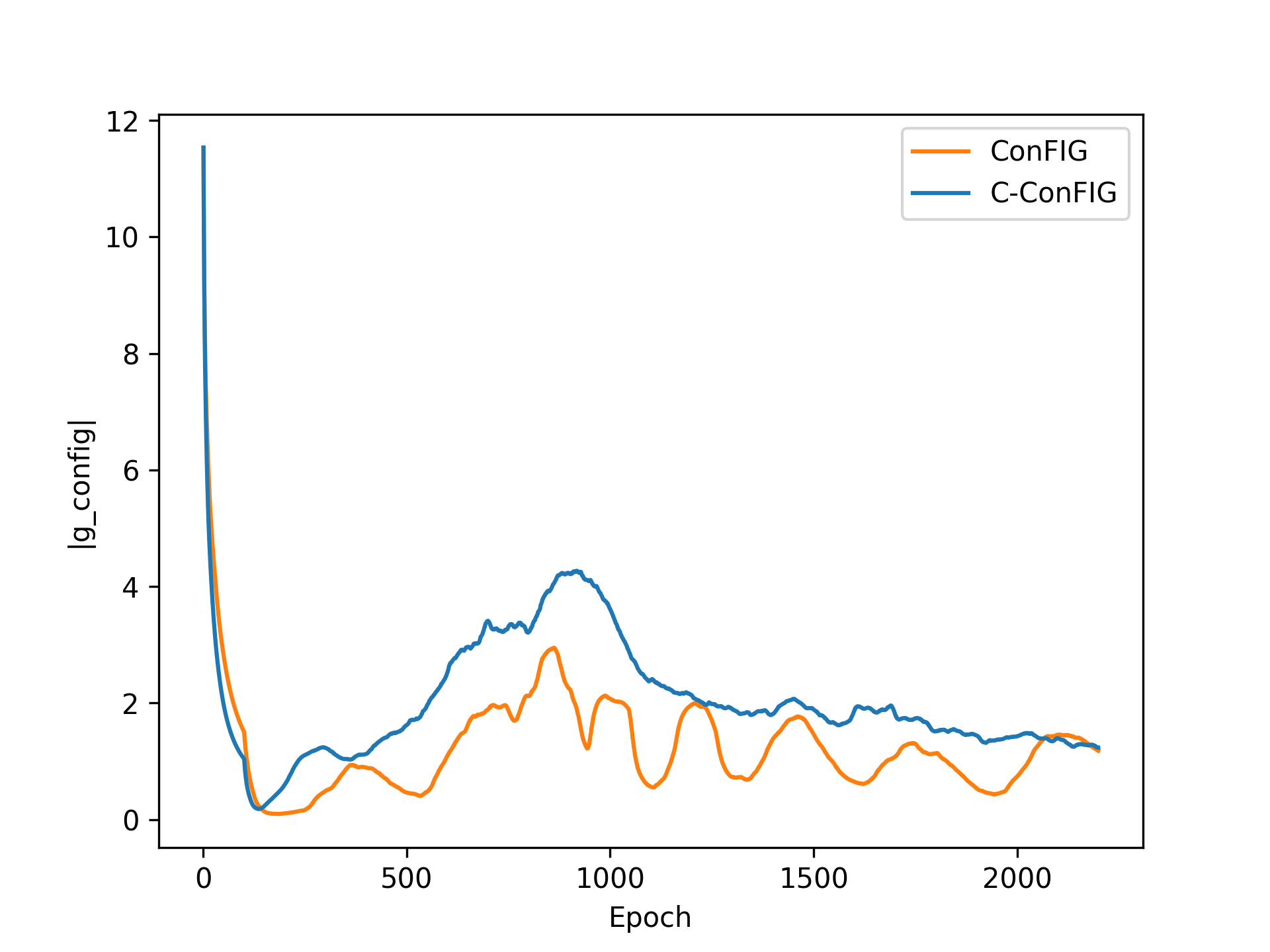}
        \caption{Kovasznay (two losses)}
    \end{subfigure}
    \hfill
    \begin{subfigure}[b]{0.32\textwidth}
        \centering
        \includegraphics[width=\linewidth]{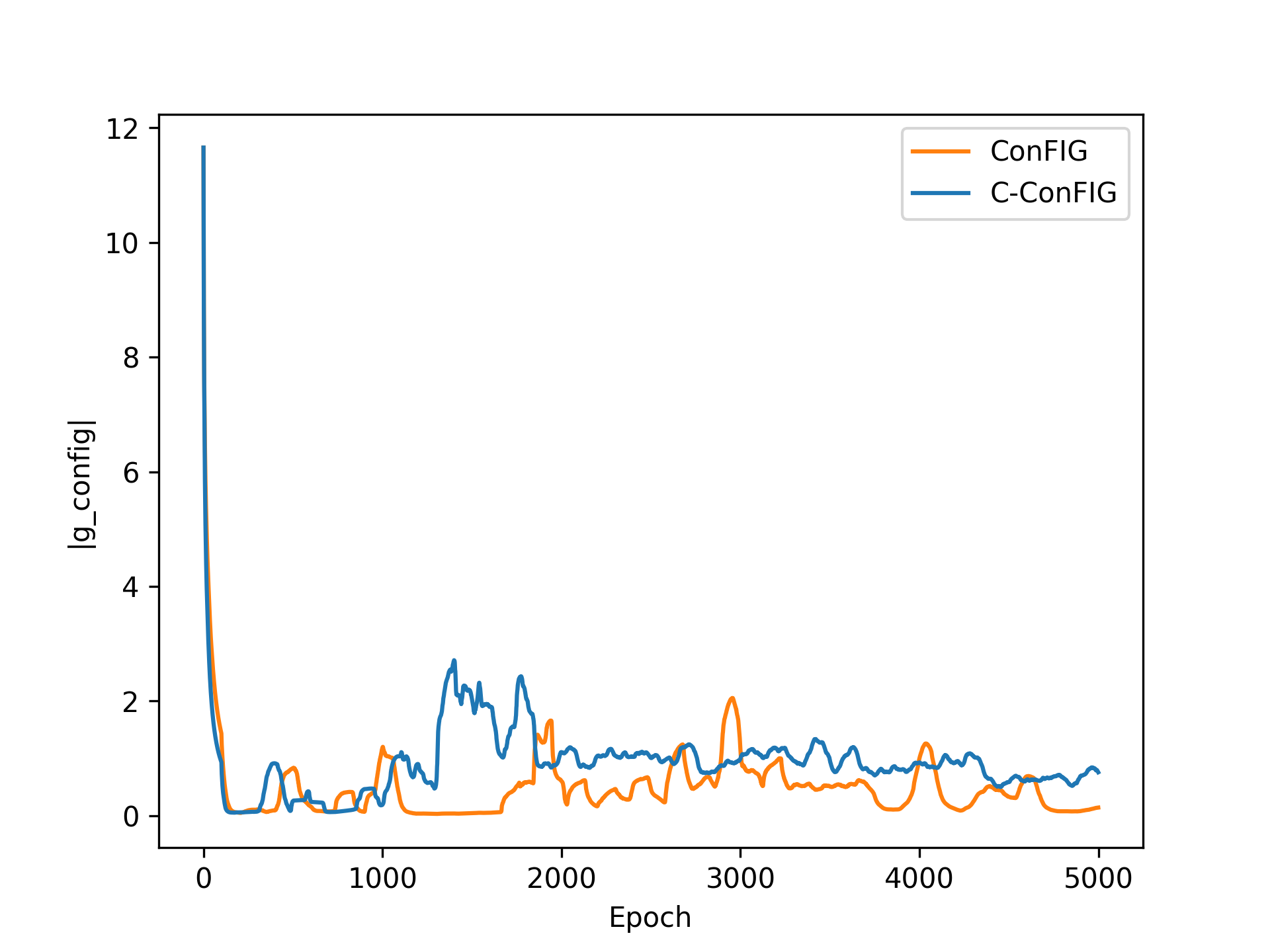}
        \caption{Kovasznay (three losses)}
    \end{subfigure}
    
    \caption{ConFIG scaling factor during early epochs}
    \label{fig:config_scaling_initial_phase}
    
\end{figure}

We identify two primary aspects in which Cautious ConFIG demonstrates improvements over the Adam + ConFIG baseline: \textbf{better distribution of update energy} and \textbf{improved directional alignment}.
We define the $L_1$ norm of the vector $u^{(t)}$ as the update energy. Cautious ConFIG skips the gradient update for parameter $i$ when $\text{sign}(g_{\text{ConFIG}})_i$ does not equal $\text{sign}(u^{(t)}_i)$.
Zeroing out a component reduces the vector's update energy. To conserve total update energy, the cautious rescaling factor $\alpha = \frac{d}{\lVert \phi \rVert_1 + 1}$ is applied to all remaining components. Consequently, update energy is redistributed from conflicting to non-conflicting coordinates. This redistribution is advantageous, as other optimizers such as RMSProp \cite{hinton2012rmsprop} and Grams \cite{cao2024grams}, whose update vector $u^{(t)}$ is constrained to the orthant of $g_{\text{ConFIG}}$, but who lack a redistribution mechanism, do not perform as well as Cautious ConFIG, as shown in Table \ref{tab:grams_rmsprop}. Grams, in particular, applies coordinatewise sign flips to mirror $u^{(t)}$ into the correct orthant. Experimental results indicate that merely aligning with the ConFIG orthant is insufficient; redistribution of update energy is also required. In coordinates where $u^{(t)}$ and $g_{\text{ConFIG}}$ disagree, the sign of $g_{\text{ConFIG}}$ should not always be trusted. In such sign conflicts, which indicate uncertainty, it is preferable to skip the update and redistribute the update energy.
The second aspect, improved directional alignment, is the primary focus of our analysis. By this, we refer to the following points:
\begin{itemize}
    \item Cautious ConFIG always increases the cosine similarity to $g_{\text{ConFIG}}$ compared to the unmodified optimizer output $u^{(t)}$ (see Proposition \ref{prop:cautious_alignment})

    \item Empirical results show that the cosine similarity between $u_{cautious}^{(t)}$ and all task gradients $g_i$ is higher than that between the unmodified output $u^{(t)}$ and $g_i$, as confirmed by Figure \ref{fig:update_task_alignment}.

    \item Cautious ConFIG produces trajectories in parameter space with fewer conflicts between different task gradients, as supported by Figure \ref{fig:burgers_kovasznay_task_conflicts_initial_phase}. This reduction in conflicts alleviates the need for conflict resolution, since conflicts between task gradients are a necessary condition for conflicts between a task gradient and a conic combination of task gradients.

    \item The Adam input $g_{\text{ConFIG}}$ exhibits greater magnitude under Cautious ConFIG compared to the baseline, due to the reduced conflicts among task gradients along Cautious ConFIG trajectories. Figure \ref{fig:config_scaling_initial_phase} provides supporting evidence.
\end{itemize}
To analyze this second aspect, we consider
a toy model that enables comparison of Cautious ConFIG and the baseline Adam + ConFIG as the number of tasks increases, optimizer lag grows, intrinsic conflict between task gradients intensifies, and the number of parameters rises.
We observe that in the first three settings, conflict resolution becomes more challenging and the baseline's ability to produce conflict-free directions declines rapidly, whereas Cautious ConFIG maintains high directional alignment. In contrast, increasing the number of parameters facilitates conflict resolution, and Cautious ConFIG's performance accelerates this improvement.

\begin{table}[htbp]
    \centering
    \begin{tabular}{lccc}
        \toprule
        \textbf{Optimizer} & \textbf{MSE}\\
        \midrule
        Cautious ConFIG & 5.775 $\pm$ 1.193\\
        Adam + ConFIG & 8.367 $\pm$ 3.482 \\
        Grams + ConFIG & 29.729 $\pm$ 20.565 \\
        RMSProp + ConFIG & 202.366 $\pm$ 96.445\\
        \bottomrule
    \end{tabular}
    \caption{Burgers Equation, All values are scaled with $10^{-4}$. Mean and standard deviation over 10 random seeds.}
    \label{tab:grams_rmsprop}
\end{table}


\subsection*{Toy Model for Directional Stability}

\subsection*{Conflict Model}
Initially, the task gradients were generated by drawing
$m$ vectors $g_i \in \mathbb{R}^d$ from a multivariate standard normal distribution $\mathcal{N}(\mathbf{0}_d, \mathbf{I}_d)$.
This model requires specification of only the hyperparameters $m$ and $d$
and is straightforward to analyze. However, when $d$ is large, as in deep learning applications, all task gradients $g_i$ become approximately pairwise orthogonal.
This scenario does not reflect the PINN setting, where significant intrinsic gradient conflicts persist even in high-dimensional spaces (see Figure \ref{fig:burgers_kovasznay_task_conflicts_initial_phase}).
Additionally, this approach does not allow for investigation of the impact of conflict strength on the method.
This consideration is important because conflicts are most pronounced at the beginning of training and diminish over time (see Figure \ref{fig:burgers_kovasznay_task_conflicts_initial_phase}). Notably, Cautious ConFIG demonstrates substantial performance improvements over ConFIG during this initial high-conflict phase (see Figure \ref{fig:cautious_config_converges_faster}), indicating the need to focus on this period.
Instead, a direction $v$ is initially sampled from a multivariate standard normal distribution
and then normalized.
This vector serves as the "conflict axis." Half of the task gradients are aligned with the conflict axis, while the other half are oriented in the opposite direction. The conflict strength $c$ linearly interpolates between the pure conflict direction (either along or opposite to the conflict axis) and a normally distributed noise vector:
\begin{align*}
    s_i &= \begin{cases}
        1, \quad i < \lfloor \frac{m}{2} \rfloor \\
        -1, \quad i \geq \lfloor \frac{m}{2} \rfloor
    \end{cases} \\
    v, z_i &\sim \mathcal{N}(\mathbf{0}_d, \mathbf{I}_d) \\
    g_i &= c \cdot s_i \cdot \frac{v}{\lVert v \rVert_2} + (1 - c) z_i
\end{align*}
The proposed model enables the representation of gradient conflicts within high-dimensional spaces.
When $c = 1$, the model exhibits maximal gradient conflicts, as half of the gradients are oriented in one direction while the remaining half are oriented oppositely.
When $c = 0$, each task gradient is sampled from a multivariate standard normal distribution, thereby recovering the simpler model described previously.
The model also reflects that, in PINNs, initial and boundary conditions typically do not exhibit significant gradient conflicts. Initial conditions are a specific type of boundary condition. In contrast, boundary conditions often display substantial gradient conflicts with the residual task gradient.
To evaluate the conflict model, the intra-alignment score $
    \mathcal{A}^{intra}(g_1, ..., g_m) = 2 \cdot \lVert \frac{1}{m} \sum_{i = 1}^m \frac{g_i}{\lVert g_i\rVert_2}\rVert_2^2 - 1$ can be calculated
as defined by \cite{wang2026gradient}-
This metric generalizes cosine similarity, which corresponds to the intra-alignment score when $m = 2$.
\begin{figure}[htbp]
    \centering
    \includegraphics[width=0.5\linewidth]{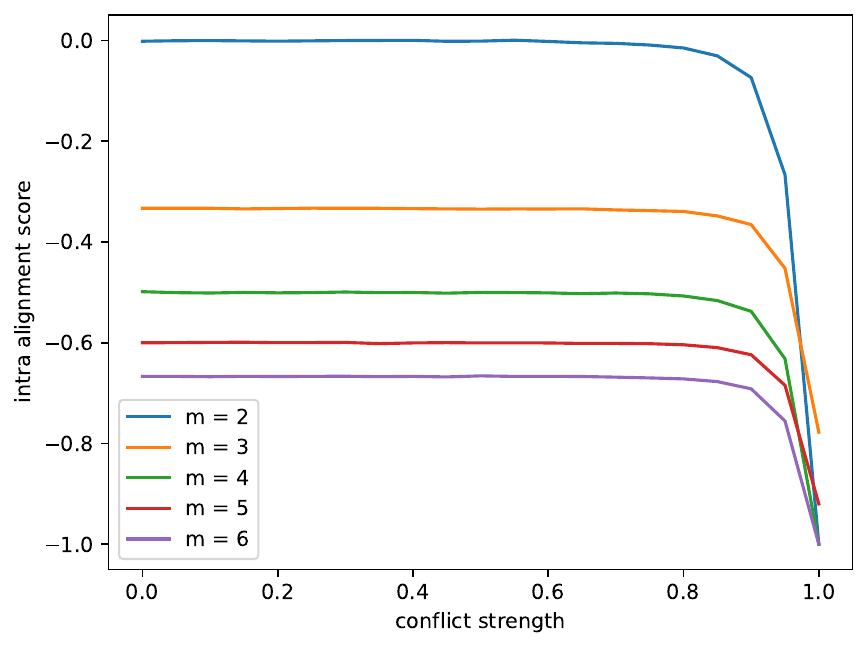}
    \caption{Impact of conflict strength and different numbers of tasks on the intra-alignment Score}
\end{figure}
The intra-alignment score remains below zero because, by construction, there are always two groups of tasks that are opposed. As the number of tasks increases, the number of conflicts also increases, resulting in a lower intra-alignment score. Additionally, the score rapidly approaches -1.0 (the lowest value) for all $m$ as the conflict strength exceeds 0.9.

\subsection*{Optimizer Model}

With the given task gradients $g_i$, we can calculate the corresponding
$g_{\text{ConFIG}}$. Then we have to compare an adaptive optimizer's output $u$ with its
cautious counterpart $u_{cautious}$. Adam e.g. has an internal state $(t, m, v)$
which is not modeled here, as including it would complicate the analysis. The directional model presented is purely static. The optimizer output is modeled as the true value $g_{\text{ConFIG}}$ plus isotropic noise $\sigma \cdot z$. Then, we can also calculate the cautious direction:
\begin{align*}
    u &= g_{config} + \sigma \cdot z,\, z \sim \mathcal{N}(\mathbf{0}_d, \mathbf{1}_d)\\
    \phi &= \mathbf{1}_{\{ u \odot g_{\text{ConFIG}} > 0  \}} \\
    u_{cautious} &= u \odot \phi
\end{align*}
The cautious scaling factor $\alpha$ can be omitted, as it affects only the magnitude of $u$, but not its direction. Since the analysis focuses on cosine similarities, and whether the resulting direction is a common descent direction (i.e., within the intersection of all conflict-free half spaces), this factor is not considered here. Our formulation can model momentum-based optimizers. When $\beta_1$ is high, the optimizer lags behind $g_{\text{ConFIG}}$, which can be represented by high $\sigma$.
However, it does not describe adaptive optimizers like Adam or RMSProp, as preconditioning with the inverse metric $M_t^{-1}$ distorts the space.

\subsection*{Qualitative Analysis}

Given $u$, $u_{cautious}$, and a set of task gradients ${g_i}$, it is possible to compute both the likelihood that the optimizer output is conflict-free and the average cosine similarity with the task gradients:
\begin{align*}
    v &\in \{ u, u_{cautious} \} \\
    P(\text{conflict-free} | v) &= \mathbb{E}_{g_1, ..., g_m}[\mathbf{1}(\bigwedge_{i} v^T g_i > 0)] \\
    \text{Avg. Cosine Sim}(v) &= \mathbb{E}_{g_i} [\text{Cos}(g_i, v)]
\end{align*}
The first quantity evaluates whether Cautious ConFIG restores the common descent property in this simplified context, which requires all dot products to be positive. In contrast, the average cosine similarity may be disproportionately influenced by outliers, such as a single task gradient with which $u_{cautious}$ has high similarity. Nevertheless, average cosine similarity remains important for quantifying the degree of gradient alignment, which is a key factor in loss reduction.
\begin{figure}[htbp]
    \centering
    \includegraphics[width=0.9\linewidth]{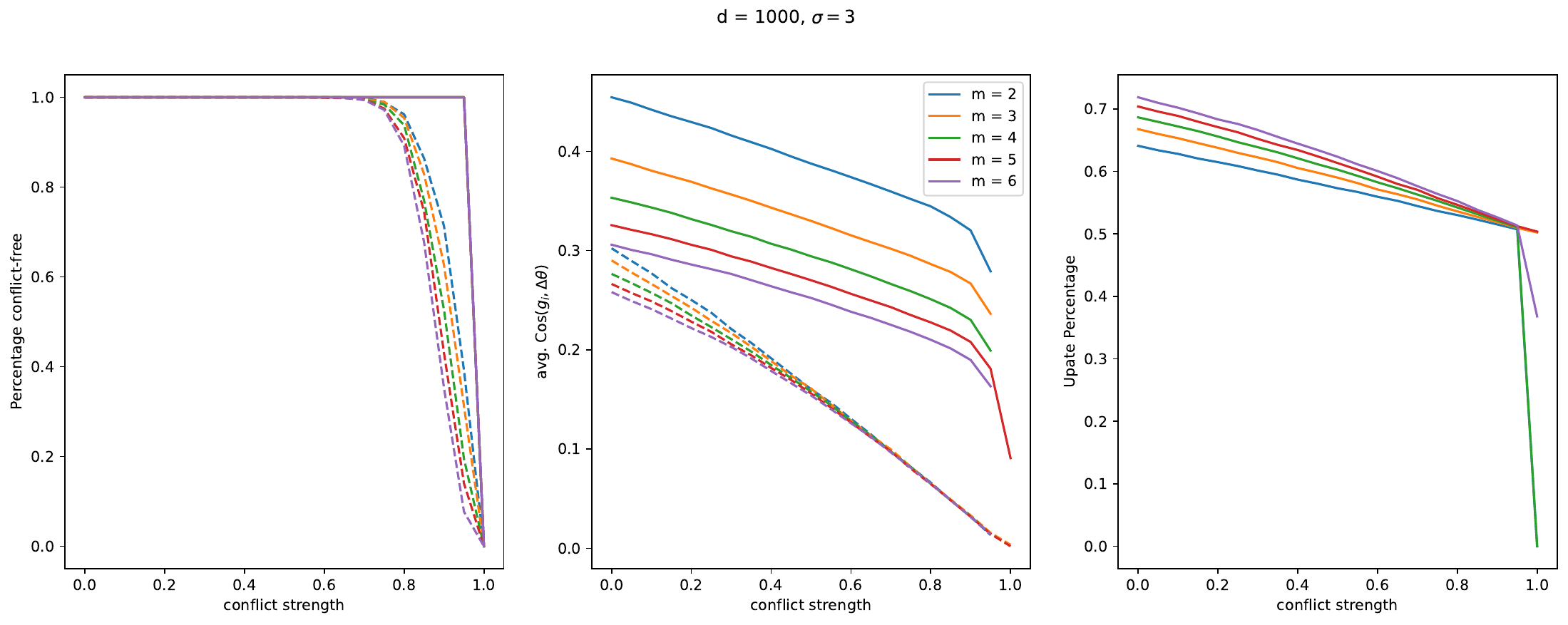}
    \caption{The effect of conflict strength at constant optimizer noise.
    Cautious ConFIG (solid), ConFIG (dashed)}
    \label{fig:constant_noise}
\end{figure}
Figure \ref{fig:constant_noise} demonstrates that as $c$ increases, the average cosine similarity between the optimizer output $u$ and the task gradients rapidly approaches zero, regardless of the number of tasks $m$. In contrast, for $u_{cautious}$, the decrease in average cosine similarity with increasing $c$ occurs at a much slower rate. Furthermore, an increase in the number of tasks results in a lower average cosine similarity, indicating that conflict resolution becomes more challenging as the number of losses increases. The probability of conflict-freeness also declines sharply at conflict strengths $c > 0.9$ for $u$, whereas it remains nearly 100\% until $c \approx 1$ for $u_{cautious}$.
\begin{figure}[htbp]
    \centering
    \includegraphics[width=0.9\linewidth]{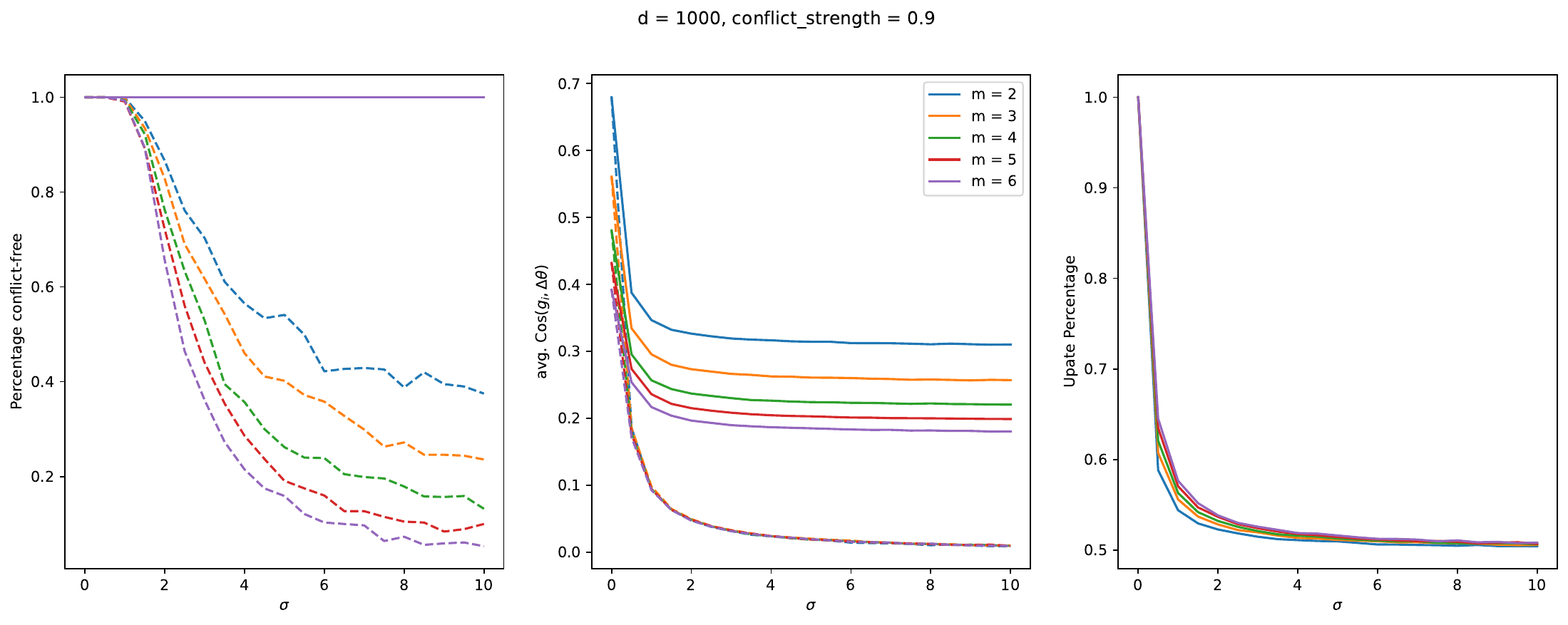}
    \caption{The effect of optimizer noise at constant conflict strength.
    Cautious ConFIG (solid), ConFIG (dashed)
    }
    \label{fig:optimizer_noise}
\end{figure}
Given the model $u = g_{\text{ConFIG}} + \sigma \cdot z$, where $z \in \mathcal{N}(0, I_d)$, increasing $\sigma$ (the optimizer noise) causes the optimizer input and output to become increasingly unaligned. This scenario can represent the behavior of a momentum-based optimizer without preconditioning as $\beta_1$ increases. Figure \ref{fig:optimizer_noise} illustrates this relationship.
As optimizer noise increases, the probability that $u$ is a descent direction rapidly decreases. Furthermore, the average cosine similarity approaches zero quickly, independent of $m$. In contrast, $u_{cautious}$ consistently exhibits a high probability of producing a common descent direction, and its average cosine similarity converges to a positive constant rather than decaying to zero. Notably, scenarios with fewer tasks display a slightly higher plateau, indicating that a reduced number of tasks facilitates conflict resolution.
\begin{figure}[htbp]
    \centering
    \includegraphics[width=0.9\linewidth]{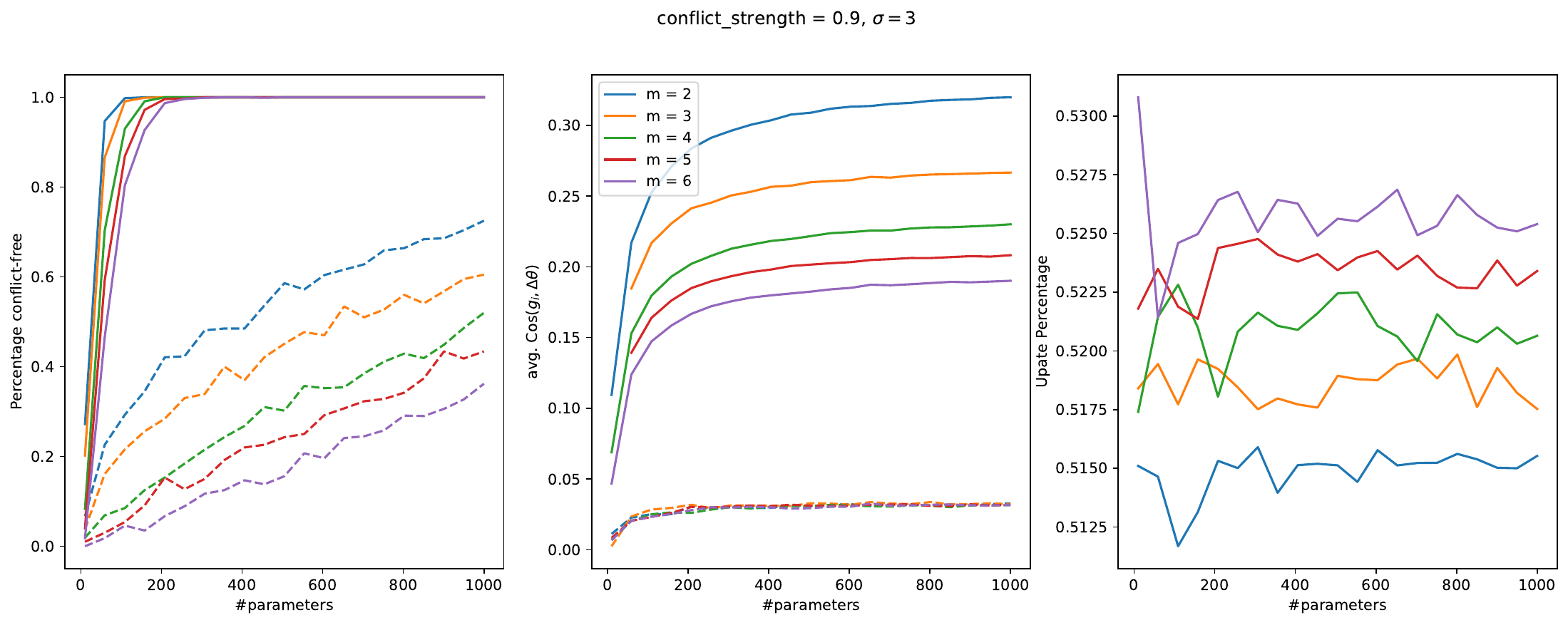}
    \caption{The effect of model size at constant conflict strength and optimizer noise.
    Cautious ConFIG (solid), ConFIG (dashed)
    }
    \label{fig:increasing_model_size}
\end{figure}
Figure \ref{fig:increasing_model_size} demonstrates that as the number of trainable parameters $d$ increases, the probability of $u$ being a conflict-free direction also increases. A high $d$, which corresponds to a larger space for the conflict-free cone, facilitates conflict resolution. Previous results indicate that a low $m$ similarly eases conflict resolution, since the conflict-free cone is defined by the intersection of half-spaces and fewer intersections occur. However, the probability of $u$ being conflict-free increases only gradually with model size, and the average cosine similarity stagnates at a low positive value as model size grows. In contrast, for $u_{cautious}$, the probability of conflict-freeness rises sharply at $d \approx 200$ and remains near 100\% for larger $d$. In this case, anchoring amplifies the advantages of high $d$. Furthermore, the average cosine similarity between $u_{cautious}$ and the task gradients plateaus at substantially higher values (exceeding 15\%), and lower $m$ results in higher plateaus. Cautious masking eliminates the degeneracy in $m$ for the average cosine similarity and significantly increases the plateau values.
Although this model greatly simplifies the situation by excluding preconditioning that distorts space and internal optimizer state, the probability of conflict-freeness and the average cosine similarity decrease rapidly when task conflicts are severe (high conflict strength $c$), the optimizer lag is large (modeled by high optimizer noise $\sigma$), the number of model parameters is small (low $d$), or the number of tasks $m$ is high. In all of these scenarios, performance degradation is significantly slowed by cautious updates until conflicts reach their maximum ($c \approx 1$). Even in this extreme case, the average cosine similarity does not vanish but instead plateaus at a positive value significantly greater than zero. A high parameter count $d$ facilitates conflict resolution, and this effect is further amplified by cautious masking.

\section{Conclusion}

In this paper, we identified that the mathematical guarantees of the ConFIG update direction are lost when this direction is used as input to an adaptive optimizer such as Adam. Consequently, the parameter update direction is no longer guaranteed to be conflict-free with respect to the task gradients.
We proposed Cautious ConFIG, a simple three-line extension to
ConFIG, which is easily parallelizable, introduces negligible computational overhead and does not require additional hyperparameters. Empirically, it restores the conflict-freeness property for the modified optimizer output prior to the parameter update. We evaluated Cautious ConFIG on three nonlinear PINN scenarios (Burgers, Schrödinger, Kovasznay)
where it demonstrates a relative improvement of 19\% to 28\% over the baseline.
Additionally, we present a toy model that demonstrates the superiority of Cautious ConFIG in task alignment under conditions of significant optimizer lag, numerous tasks, and pronounced gradient conflicts.

Promising future research directions include integrating cautious updates with other conflict-resolution algorithms such as PCGrad or Nash-MTL, which are also frequently used in conjunction with adaptive optimizers.In addition, it will be highly interesting to incorporate the method into physics-constrained generative models \cite{baldan2026pbfm} and hybrid numerical methods \cite{franz2025pict}.
On the theoretical side, the toy model we provide could be analyzed mathematically to derive relationships involving relevant quantities such as task number and conflict strength, as our current work is limited to computational experiments. Additionally, we
could develop an improved toy system that incorporates both strong conflicts in high-dimensional spaces, as observed in the PINN setting, and a more realistic yet analytically tractable optimizer model.

\bibliography{iclr2027_conference}
\bibliographystyle{iclr2027_conference}

\appendix
\section{Appendix}

\subsection*{Proposition (Cautious Masking Increases Cosine Similarity with ConFIG direction)}\label{prop:cautious_alignment}

 It is demonstrated that the cautious direction $u_{cautious}$ is always more aligned with $g_{\text{Config}}$ than its unprocessed counterpart:
\begin{align*}
\text{Cos}(g_{\text{ConFIG}}, u_{cautious}) \geq \text{Cos}(g_{\text{ConFIG}}, u)
\end{align*}
The cosine similarity does not depend on the scaling of the arguments. Since $u_{cautious} \propto \phi \odot u$, we have:
\begin{align*}
\text{Cos}(g_{\text{ConFIG}}, u_{cautious}) &\geq \text{Cos}(g_{\text{ConFIG}}, u) \\
\text{Cos}(g_{\text{ConFIG}}, \phi \odot u) &\geq \text{Cos}(g_{\text{ConFIG}}, u) \\
\frac{g_{\text{ConFIG}}^T (\phi \odot u)}{\lVert g_{\text{ConFIG}} \rVert_2 \lVert \phi \odot u  \rVert_2} &\geq \frac{g_{\text{ConFIG}}^T u}{\lVert g_{\text{ConFIG}}\rVert_2 \lVert u \rVert_2}\\
\frac{g_{\text{ConFIG}}^T (\phi \odot u)}{\lVert \phi \odot u \rVert_2} &\geq \frac{g_{\text{ConFIG}}^T u}{\lVert u \rVert_2}
\end{align*}
We have to prove the last relation. We see that:
\begin{align*}
g_{config}^T (\phi \odot u) &= \sum_{j = 1}^d (g_{config})_j (\phi \odot u)_j \\
&= \sum_{j = 1}^d (g_{\text{ConFIG}})_j \, u_j \mathbf{1}\{(g_{\text{ConFIG}})_j \cdot u_j > 0 \} \\
& \geq\sum_{j = 1}^d (g_{\text{ConFIG}})_j u_j \\
&= g_{\text{ConFIG}}^T u
\end{align*}
Furthermore, $\lVert \phi \odot u \rVert_2 \leq \lVert u \rVert_2$ because setting a component to zero can only decrease the vector's length. Therefore, the numerator increases while the denominator decreases, confirming that the relation holds.

\subsection*{Proposition (Cautious Masking creates non-negative dot products)}\label{prop:cautious_pos_dot_prod}

Consider the case where $\phi = \mathbf{1}\{ u \odot g > 0 \}$. It follows that $\langle u \odot \phi, g \rangle \geq 0$:
\begin{align*}
    \langle u \odot \phi, g \rangle &= \sum_{j = 1}^d (u \odot \phi)_j g_j \\
    &= \sum_{j = 1}^d u_j \; g_j \; \mathbf{1}\{\text{sign}(u_j) = \text{sign}(g_j)\}\\
    &\geq \sum_{j = 1}^d 0\\
    &\geq 0
\end{align*}

\subsection*{Experimental Details}\label{app:experimental details}

\textbf{The one-dimensional viscous Burgers’ equation} is considered in the following form:
\begin{align*}
    \frac{\partial u}{\partial t} + u \frac{\partial u}{\partial x} = \nu \frac{\partial^2 u}{\partial x^2}
\end{align*}
The objective is to find the solution $u: [-1, 1] \times [0, 1] \rightarrow \mathbb{R}$
This function models the velocity of a one-dimensional fluid, such as fluid flow within a pipe.
The initial condition is defined as follows:
\begin{align*}
    u(x, 0) = -\sin(\pi x), \quad x \in [-1, 1]
\end{align*}
The boundary conditions are specified as follows:
\begin{align*}
    u(1, t) = u(-1, t) = 0, \quad t \in [0, 1]
\end{align*}
The viscosity of the fluid is set to $\nu = \frac{0.01}{\pi}$.

\textbf{The one-dimensional nonlinear Schrödinger equation} is defined as follows:
\begin{align*}
i \frac{\partial h}{\partial t} + \frac{1}{2} \frac{\partial^2 h}{\partial x^2}
+ |h|^2 h = 0
\end{align*}
The computational domain is defined as $x \in [-5, 5]$ and $t \in [0, \frac{\pi}{2}]$.
Periodic boundary conditions are imposed, along with the initial condition
$h(x, 0) = 2 \text{sech}(x)$.

The partial differential of the \textbf{Kovasznay flow} is given by:
\begin{align*}
    (u \cdot \nabla) u &= -\nabla p + \frac{1}{Re} \nabla^2 u \\
    \nabla \cdot u &= 0
\end{align*}
Our domain is:
\begin{align*}
    x &\in [-0.5, 1] \\
    y &\in [-0.5, 1.5]
\end{align*}
The analytical solution (used for the Dirichlet boundary condition) is:
\begin{align*}
    u_x(x, y) &= 1 - e^{\lambda x} \cos(2 \pi y) \\
    u_y(x, y) &= \frac{\lambda}{2 \pi} e^{\lambda x} \sin (2 \pi y) \\
    p(x, y) &= \frac{1}{2} (1 - e^{2 \lambda x})
\end{align*}
Here, we defined:
\begin{align*}
    \nu &= \frac{1}{Re} = \frac{1}{40} \\
    \lambda &= \frac{1}{2 \nu} - \sqrt{\frac{1}{4 \nu^2} + 4 \pi^2 } 
\end{align*}

\begin{table}[H]
    \centering
    \caption{Shared Hyperparameters}
    \label{tab:shared_hyperparameters}
    \begin{tabular}{lc}
        \toprule
        \textbf{Parameter} & \textbf{Value} \\
        \midrule
        Learning Rate Scheduler & Cosine \\
        Learning Rate Max/Min & 1e-3/ 1e-4 \\
        Hidden Dimensions & [50, 50, 50, 50, 50] \\
        Activation Function & Tanh \\
        Initialization & Xavier \\
        use\_least\_squares (ConFIG) & False \\
        Sampling Method & Latin Hypercube \\
        Seeds & 0 ... 9 \\
        \bottomrule
    \end{tabular}
\end{table}

\begin{table}[H]
    \centering
    \caption{PINN-specific Hyperparameters}
    \label{tab:specific_hyperparameters}
    \begin{tabular}{lcccc}
        \toprule
        \textbf{PINN} & \textbf{Epochs} & \textbf{Initial Points} & \textbf{Boundary Points} & \textbf{Interior Points}\\
        \midrule
        Burgers & 5000 & 50 & 50 & 2000 \\
        Schrödinger & 100 000 & 500 & 500 & 20 000 \\
        Kovasznay & 10 000 & - & 500 & 20 000 \\
        \bottomrule
    \end{tabular}
\end{table}

\end{document}